\documentclass[journal]{IEEEtran}
\usepackage{amsmath,amsfonts}
\usepackage{algorithmic}
\usepackage{algorithm}
\usepackage{array}
\usepackage[caption=false,font=normalsize,labelfont=sf,textfont=sf]{subfig}
\usepackage{textcomp}
\usepackage{stfloats}
\usepackage{booktabs}
\usepackage{url}
\usepackage{times}
\usepackage{verbatim}
\usepackage{graphicx}
\usepackage{makecell}
\usepackage{cite}
\usepackage{multirow}
\usepackage{amsmath}
\usepackage{bbding}
\usepackage{subfloat}

\begin{document}
	
	\title{Multimodal Spatio-temporal Graph Learning for Alignment-free RGBT Video Object Detection}
	
	\author{Qishun Wang, Zhengzheng Tu, Chenglong Li, Bo Jiang}
	
	\markboth{Journal of \LaTeX\ Class Files,~Vol.~14, No.~8, August~2021}%
	{Shell \MakeLowercase{\textit{et al.}}: A Sample Article Using IEEEtran.cls for IEEE Journals}

	\maketitle
	
	\begin{abstract}
		RGB-Thermal Video Object Detection (RGBT VOD) can address the limitation of traditional RGB-based VOD in challenging lighting conditions, making it more practical and effective in many applications.
		However, similar to most RGBT fusion tasks, it still mainly relies on manually aligned multimodal image pairs. 
		In this paper, we propose a novel  Multimodal Spatio-temporal Graph learning Network (MSGNet) for alignment-free RGBT VOD problem by leveraging the robust graph representation learning model. 
		Specifically, we first design an Adaptive Partitioning Layer (APL) to estimate the corresponding regions of the Thermal image within the RGB image (high-resolution), achieving a preliminary inexact alignment. 
		Then, we introduce the Spatial Sparse Graph Learning Module (S-SGLM) which employs a sparse information passing mechanism on the estimated inexact alignment to achieve reliable information interaction between different modalities. 
		Moreover, to fully exploit the temporal cues for RGBT VOD problem, we introduce Hybrid Structured Temporal Modeling (HSTM), which involves a Temporal Sparse Graph Learning Module (T-SGLM) and Temporal Star Block (TSB). T-SGLM aims to filter out some redundant information between adjacent frames by employing the sparse aggregation mechanism on the temporal graph. Meanwhile, TSB is dedicated to achieving the complementary learning of local spatial relationships. 
		Extensive comparative experiments conducted on both the aligned dataset VT-VOD50 and the unaligned dataset UVT-VOD2024 demonstrate the effectiveness and superiority of our proposed method. Our project will be made available on our website for free public access.
	\end{abstract}

\section{Introduction}
\label{sec:intro}
\begin{figure}[t]  
	\centering
	\includegraphics[width=0.48\textwidth]{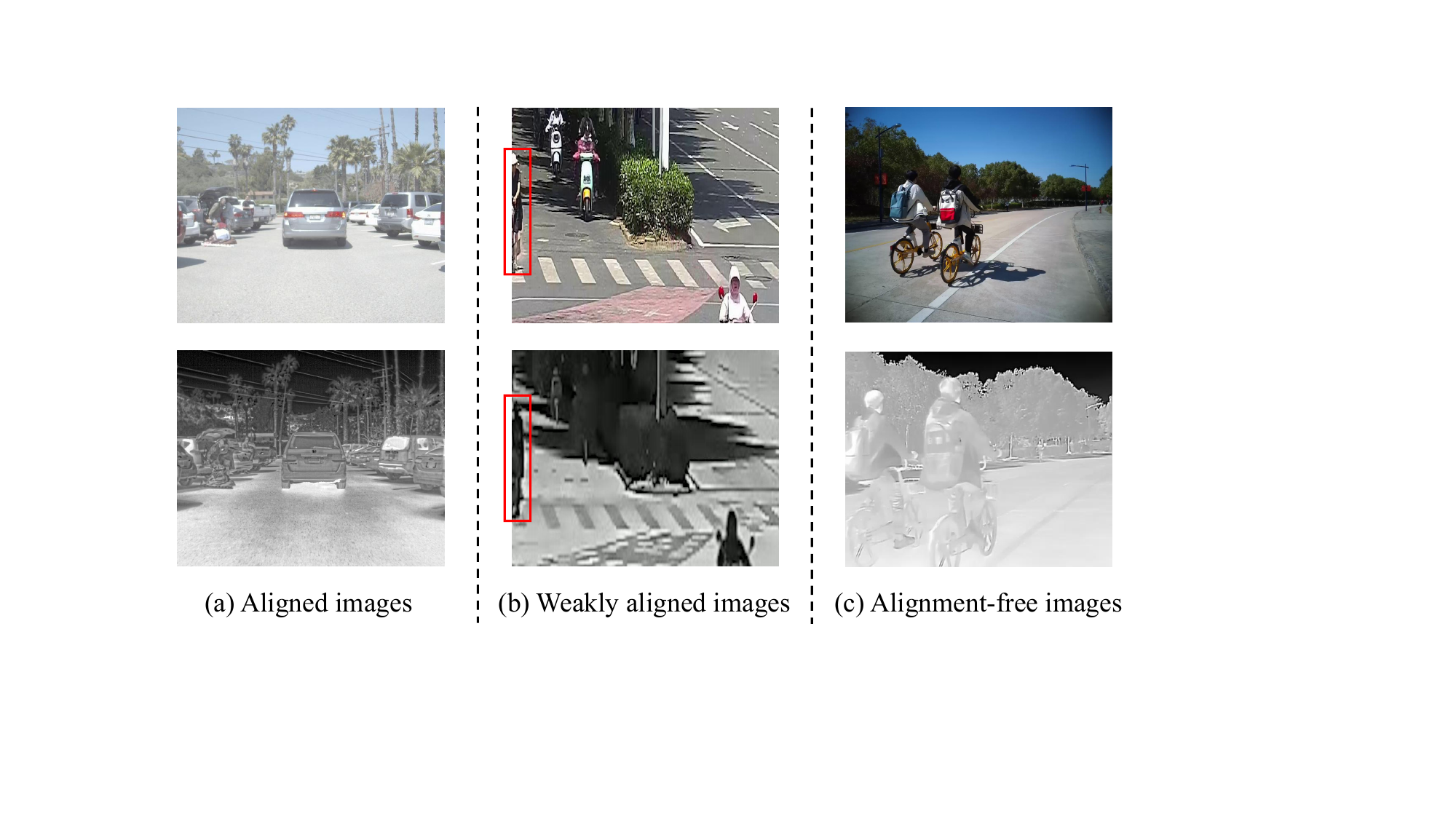}
	\caption{Owing to disparities in sensor field of view and resolution, RGBT image pairs directly captured by sensors in real-world scenarios typically resemble the situation depicted in (c), where the two modalities are not strictly spatially aligned. Previous studies often necessitate spatially aligned or near-aligned image pairs, as shown in (a) and (b), as input. However, achieving such alignment typically requires manual intervention, involving the use of calibration data or specialized algorithms for preprocessing, which can be cumbersome and impractical for specific applications.} 
	\label{fig1}	
\end{figure}

\IEEEPARstart{T}{he} emergence of RGB-Thermal (RGBT) Video Object Detection (VOD) \cite{tu2023erasure} overcomes the limitations of RGB-based VOD in harsh lighting conditions which is more important in practical applications. 
One intuitive way to address this problem is to employ RGBT fusion techniques, and some fusion methods have been studied in recent works \cite{xiang2022rgb,peng2023hafnet,kim2024causal}. 
For example, 
EINet \cite{tu2023erasure} utilizes the negative activation features of Thermal images to suppress noise in RGB ones, thereby achieving  RGBT fusion. PTMNet \cite{wang2025high} leverages the difference tensor between RGBT modalities to capture their complementary features. 
However, they all require aligned RGBT image pairs, as depicted in Fig. \ref{fig1} (a) for training input, which consumes considerable manpower and resources. 
Therefore, some RGBT fusion methods have explored the usage of weakly aligned image pairs as shown in Fig. \ref{fig1} (b) for training to achieve more robust fusion models. 
For example, Sun \emph{et al}. \cite{sun2022drone} propose assigning uncertainty weights to unaligned instances based on the Intersection over Union (IoU) values of the object positions in the two modalities. This approach equips the detection algorithm with an uncertainty score, enabling it to perform detection under conditions of uncertainty perception.
Chen \emph{et al}. \cite{chen2024weakly} propose a cross-modality spatial offset modeling module to estimate the spatial distribution offset between two modalities, thereby implicitly determining the optimal position for fusion.

However, in many real-world scenarios, RGBT image pairs may be unaligned due to differences in parameters such as sensing wavelength and focal length across different spectral sensors, which lead to variations in resolution and viewing angle in multimodal imaging, as illustrated in Fig. \ref{fig1} (c). 
To address this issue, some recent studies have begun to explore the usage of unaligned RGBT image pairs. 
For example, Wang \emph{et al}. \cite{wang2024alignment} propose a semantics-guided asymmetric correlation network to learn feature correspondences between unaligned salient regions across modalities. Liu \emph{et al}. \cite{liu2024non} develop a temporal-iterated homography estimation module for performing cross-modal spatial alignment, enabling adaptation to both weakly aligned and unaligned data scenarios. Similarly, Song \emph{et al}. \cite{song2024misaligned} provide an alignment network to estimate the deformation field from RGB to Thermal images, aiming to eliminate spatial scene distribution bias.

Upon reviewing the existing literature, we identify several limitations that generally characterize the current state of research in this domain. First, most existing methods rely on implicit alignment strategies, which are insufficient to address the significant misalignment of objects in RGBT image pairs. Second, previous approaches based on matrix estimation often encounter difficulties in ensuring the stability of the transformation, primarily due to the semantic disparities between RGB and Thermal modalities. Consequently, the development of alignment-free RGBT VOD remains a challenging yet critically important task that warrants further attention.

To address these challenges, in this paper, we propose a novel framework for alignment-free RGBT VOD by leveraging robust graph representation learning. Specifically, we first introduce a new Adaptive Partitioning Layer (APL) that estimates a scale factor and identifies a rectangular region to roughly determine the spatial correspondence between the Thermal image and the high-resolution RGB image. This initial alignment step provides a foundation for subsequent feature fusion while mitigating the impact of significant misalignment. Next, to robustly integrate the complementary features of the roughly aligned regions, we develop a Spatial Sparse Graph Learning Module (S-SGLM). In this module, two sets of nodes are constructed, representing feature values extracted from the RGB and Thermal images, respectively. The edges between these nodes capture the correlations between corresponding features across the two modalities. By adopting a sparse message aggregation mechanism, S-SGLM facilitates the aggregation of highly correlated patches, thereby alleviating the issue of inexact alignment between different modalities. This approach enhances the robustness of feature fusion and improves the overall detection performance in the absence of strict alignment.

Furthermore, to effectively utilize temporal cues for the RGBT VOD problem, we introduce a new Hybrid Structured Temporal Modeling (HSTM) module. 
Specifically, in HSTM, we first 
build a temporal graph to encode the temporal relations between patches of adjacent frames.
Then, we employ a Temporal Sparse Graph Learning Module (T-SGLM) to eliminate the redundant temporal information by adaptively exploiting the sparse interactions between patches across different frames. 
Finally, a Temporal Star Block (TSB) is further proposed to learn the local spatial structure consistency over temporal.
HSTM prevents the instability encountered in region proposal-based methods during interactions in intricate object scenarios. Also, it circumvents the significant redundancy associated with feature map-based methods in temporal feature aggregation. 

Incorporating the aforementioned enhancements, we introduce a Multimodal Spatio-temporal Graph learning Network (MSGNet) as a solution for the alignment-free RGBT VOD problem. Overall, the main contributions of this paper are summarized as follows:

\begin{itemize}
	\item We propose APL to flexibly discern the spatial information distribution discrepancy in RGBT image pairs and achieve coarse-grained alignment.
	\item We propose S-SGLM for constructing sparse multimodal graphs with weakly aligned RGBT image pairs, leveraging highly correlated multimodal features for informed fusion.
	\item We propose HSTM to eliminate temporal redundancy while preserving the capacity to learn local spatial structural consistency.
	\item We propose MSGNet for alignment-free RGBT VOD, which demonstrates top-tier performance on both the aligned VT-VOD50 and unaligned UVT-VOD2024 benchmarks, setting a new state-of-the-art (SOTA) standard.
\end{itemize}

\begin{figure*}[t]  
	\centering
	\includegraphics[width=0.85\textwidth]{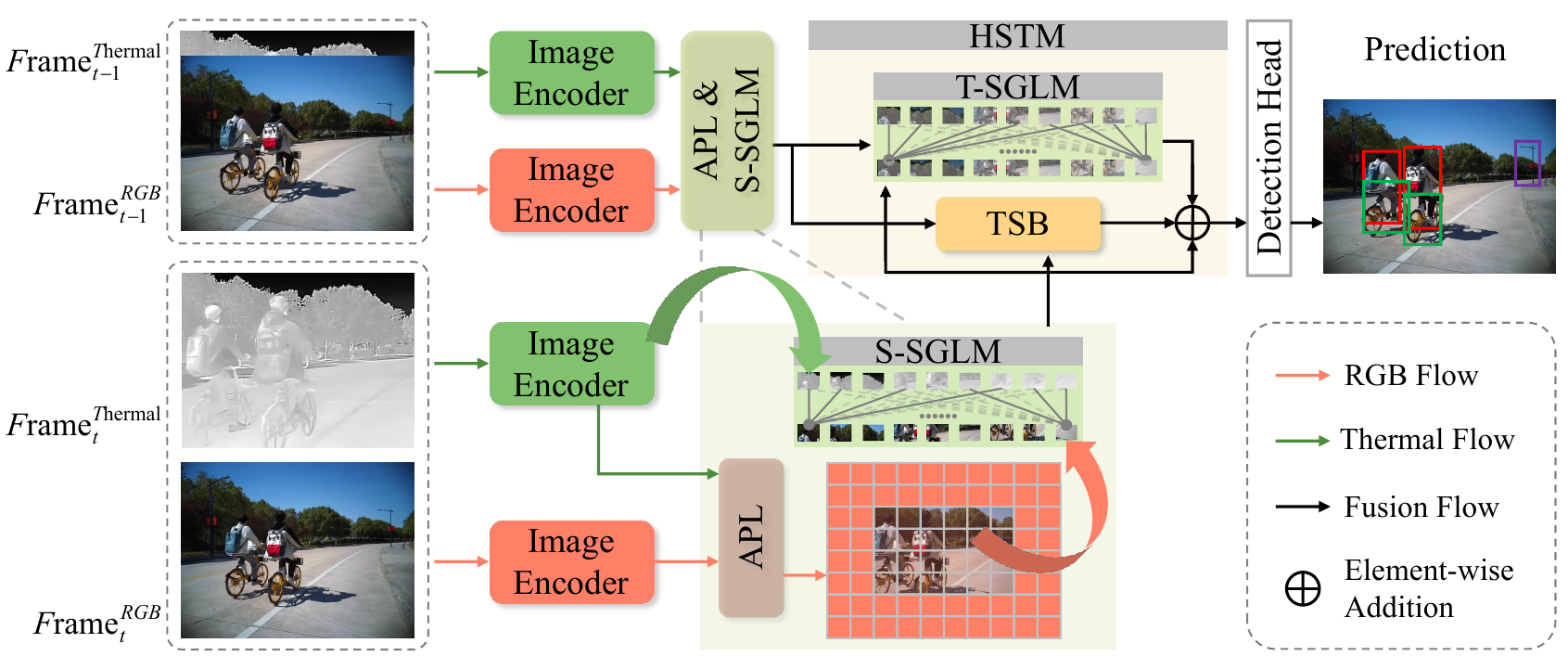}
	\caption{The schematic diagram of MSGNet. We utilize a total of four input frames. MSGNet initially employs APL and S-SGLM to address the challenge of alignment-free fusion across disparate modalities. Subsequently, HSTM is leveraged to encapsulate the temporal features at two distinct time points. Ultimately, the detection head processes these enriched features to yield the final prediction.
	} 
	\label{net}	
\end{figure*}

\section{Related Work}
In this section, we introduce and analyze studies on the fusion of RGBT images without strict alignment, as well as the significant advancements in VOD technologies.

\subsection{RGBT Fusion without Strict Registration}
Existing RGBT image pairs frequently necessitate extensive manual processing to mitigate differences in original spatial distribution, which diverges from reality and impedes the development of large-scale datasets. Li \emph{et al}. \cite{li2024deep} propose a dual-branch relation prediction network to capture spatial correspondence between unaligned feature maps and subsequently perform dynamic aggregation of these features. However, this method still relies on registered RGBT image pairs to inject prior knowledge into the training module, and unaligned RGBT images here exhibit weak alignment scenarios that are more idealized than real imaging conditions. Wang \emph{et al}. \cite{wang2024alignment} introduce Semantics-guided Asymmetric Correlation Network (SACNet) to identify foreground areas of interest in both modalities, followed by feature fusion in these local areas to enhance robust unaligned RGBT salient object detection. Nevertheless, when multiple objects are present in the image, the network’s focus may become dispersed, resulting in diminished fusion effectiveness. 

Liu \emph{et al}. \cite{liu2024non} propose using homography matrix estimation to learn the spatial mapping relationship between unaligned RGBT image pairs, followed by spatial transformation based on this matrix to achieve alignment. While this approach is simple and effective, it cannot overlook the impact of modality differences and the low quality of single-modality data on homography matrix estimation. Zhang \emph{et al}. \cite{zhang2024amnet} develop a Mutual-interacted Spatial Alignment (MSA) module for RGBT tracking to address positional deviations of objects in images from different modalities. However, the data processed by MSA is aligned at the image level but not at the instance level, meaning the positions of the tracked objects in the two modalities exhibit some overlap. In a comparable data scenario, Zhang \emph{et al}. \cite{zhang2019weakly} propose utilizing the feature proposals of the object in different modalities to apply spatial offset and fusion for multispectral object detection, contingent upon limited positional distribution deviation between them.

Consequently, the above methods may not be robust when confronted with completely unaligned image pairs at both the pixel and image levels.

\subsection{Video Object Detection}
As the demand for technologies such as surveillance, security, and autonomous driving grows, new challenges and opportunities arise for the development of VOD. Xu \emph{et al}. \cite{xu2022multilevel} conduct full-link interactions including frame-level, proposal-level and instance-level to solve the problem of object appearance deterioration in VOD. This contextual similarity can not only enhance the features of the object, but also update the reference features. In particular, they also propose a deformable feature alignment module to achieve more accurate pixel-level spatial alignment. The DiffusionVID \cite{roh2023diffusionvid} suggests adding random noise boxes in multiple consecutive frames, followed by applying the concept of the diffusion model to cascade denoising and enhance the prediction boxes. This method circumvents redundant feature interactions at the proposal level, thereby achieving greater efficiency. Wang \emph{et al}. \cite{wang2022ptseformer} propose the PTSEFormer, which aims to decouple the temporal and spatial features between adjacent frames and progressively model these two types of information. This approach effectively leverages the spatiotemporal consistency present in the data. To enhance the accuracy of small object detection in videos, Xiao \emph{et al}. \cite{xiao2023lstfe} integrate contextual information from long-term reference frames with the consistency features of short-term adjacent frames to align spatiotemporal features. However, this inevitably increases the computational overhead of the algorithm. How can the features of the reference frame be utilized more efficiently? The Impression Network \cite{hetang2023impression} proposes an iterative method for aggregating sparse key frame features to achieve long-range multi-frame fusion. By abandoning the use of dense continuous frames, this approach has improved reasoning speed to some extent. MaskVD \cite{sarkar2024maskvd} significantly reduces computation and latency by masking portions of the video frame, without compromising network performance, thanks to the retention of semantic information and spatiotemporal correlations between frames. The Hybrid Multi-Attention Transformer (HyMAT) module \cite{moorthy2025hybrid} is proposed to use the appearance and position of the object simultaneously to enhance relevant features, which can suppress the erroneous information interaction between objects with similar appearance to a certain extent, thereby improving detection accuracy. However, it cannot eliminate object matching errors and redundant information enhancement in complex multi-object scenes.

Yin \emph{et al}. \cite{yin2021graph} propose a Grid Message Passing Network (GMPNet), which treats each grid as a node and then constructs a k-NN graph with adjacent grids. The features of the current grid are updated through the relationship between adjacent nodes, and the information of distant grids is iteratively aggregated. However, there is not much consideration here on how to improve the efficiency of the network when iteratively aggregating information.

\begin{figure*}[t]  
	\centering
	\includegraphics[width=0.8\textwidth]{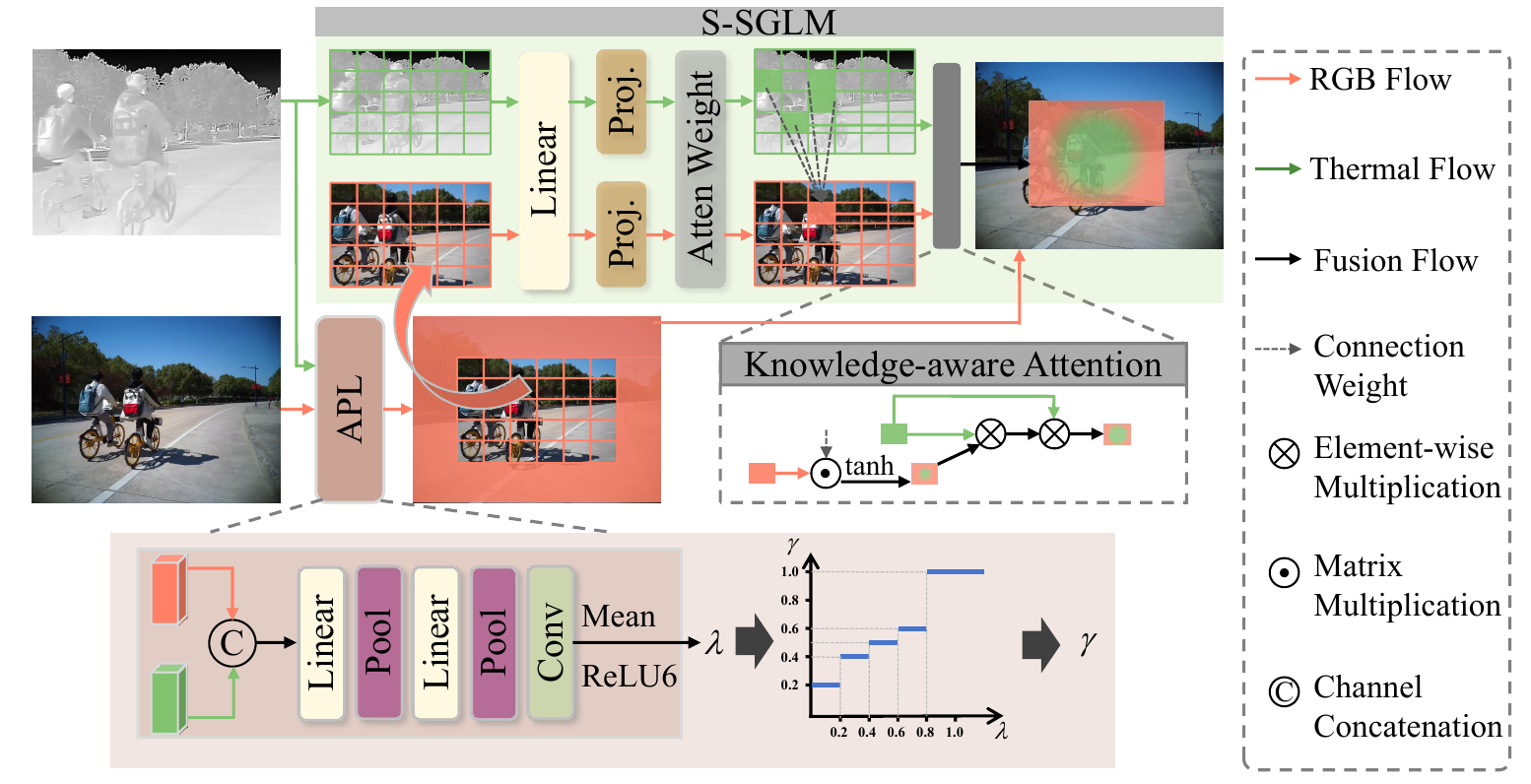}
	\caption{The structure of APL and S-SGLM. The APL module first jointly predicts the feature maps of RGB and Thermal images. To achieve this, we employ a progressive approach to derive a value $\lambda$, which serves as an intermediate representation. Subsequently, we apply our predefined rules to compute a value $\gamma$ based on $\lambda$. The value $\gamma$ is utilized to identify and select a local region within the RGB image that corresponds to a larger field of view. This region is hypothesized to spatially align with the area of interest in the thermal image. The subsequent interaction between the RGB and thermal image features is then facilitated through the composition and aggregation of information from these two corresponding regions. This process enables the model to effectively integrate multimodal data while preserving spatial coherence between the RGB and thermal domains. Among them, ``Proj.'' denotes the projection layer implemented with a convolutional kernel of size \(1 \times 1\).} 
	\label{rgbtfusion}	
\end{figure*}

\section{Method}

We select the advanced one-stage detector YOLOV8 \cite{Jocher_Ultralytics_YOLO_2023} as our base detector due to its performance and efficiency, as well as its capability for end-to-end training. To incorporate multimodal temporal information, we input multiple frames and extend the network branches. The image encoder and detection head align with the baseline design.

\subsection{Image Encoder}
\begin{figure}[htbp]  
	\centering
	\includegraphics[width=0.48\textwidth]{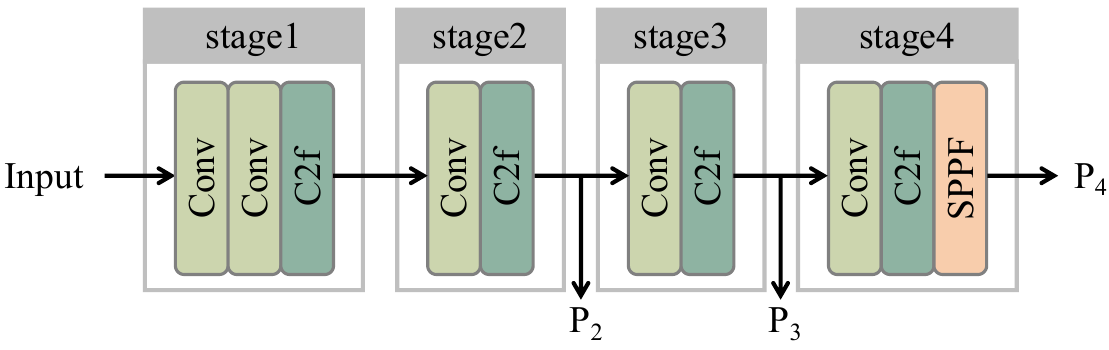}
	\caption{Schematic diagram of the image encoder used by our method. The encoder accepts the input image and produces three layers of feature maps with distinct scales. They have both shallow detail information and high-level semantic information. These feature maps are subsequently transmitted to the intermediate stage of the network for feature fusion. } 
	\label{backbone}	
\end{figure}

In Fig. \ref{backbone}, we stack $Frame^{Thermal}_{t-1}$ and $Frame^{RGB}_{t-1}$ for a more beautiful composition. The semantic encoders we equip for the four input images are completely symmetrical in structure, and they share parameters with each other. This is because we need to ensure that the learning of image encoders in different modalities does not interfere with each other and can fully extract the semantic features within a single modality.

The basic unit of the image encoder consists of convolutional layers, C2f layers, and the SPPF layer, please refer to the baseline \cite{Jocher_Ultralytics_YOLO_2023} for more details about them. The image encoder outputs three layers of multi-scale features, $P_{2}$-$P_{4}$. In the middle stage and head of the network, the three feature maps are involved in learning at different scales, although the learning paths and methods are completely symmetrical. Therefore, we only show single-stream information in Fig. \ref{net} to ensure the clarity of the network diagram.

\subsection{APL: Adaptive Partitioning Layer}
We introduce APL to mitigate the scale misalignment issues between objects and scenes in RGBT image pairs. APL specifically concatenates the two feature maps along the channel dimension, and through a series of linear, pool, and convolutional layers, iteratively refines the prediction to determine a scale factor $\lambda$. This factor $\lambda$ signifies the network's preliminary assessment of the scale discrepancy between the unaligned images. Utilizing the continuous value $\lambda$, we then derive the final cropping factor $\gamma$ through established rules, as depicted in Fig. \ref{rgbtfusion}. Essentially, this step functions as a routing mechanism that identifies the most suitable corresponding area for subsequent image composition.

\subsection{S-SGLM: Spatial Sparse Graph Learning Module}
Graph representation learning has recently been employed in image processing to capture long-range contextual dependencies \cite{li2024dynamic}, which can adapt to varying spatial structures. RGBT image pairs with different misalignments lack strict correspondence or regularity in spatial distribution, motivating us to use graph representation learning to aggregate features within structurally similar contexts.

The structure diagram of S-SGLM is shown in Fig. \ref{rgbtfusion}. Following the acquisition of $\gamma$ from APL, we crop the higher-resolution RGB image in a centered manner while maintaining proportionality. This cropped section of the feature map captures the common spatial distribution that the network has learned from the two modalities. Next, we resize the thermal image to match this size and utilize a linear layer with shared parameters to execute unified feature mapping. This step aims to reduce the semantic distribution disparity between the two modal features, thereby facilitating graph construction. We proceed by embedding features from both modalities and computing dense attention scores between them, which indicate the degree of information correlation between the thermal and RGB feature values. Subsequently, we create a dense connection graph using the attention scores as edge weights between each pair of feature points. However, it is important to note that dense edges are not universally essential; not all thermal feature values contribute significantly to the feature values of the current spatial position in RGB. To address this, we eliminate insignificant node pairs by selecting the top k and removing connection edges with weights below a specified threshold.

Upon completion of the aforementioned process, we obtain a sparse graph that connects the two modalities. For each point on the RGB feature map, we identify a set of feature points from the thermal image that are strongly correlated with it, as indicated by the connection edges. Ultimately, we employ a knowledge-aware attention mechanism to aggregate the high-contribution features from these thermal images into the RGB domain, thereby achieving robust feature complementation.

\subsection{HSTM: Hybrid Structured Temporal Modeling}
We have crafted the HSTM module for the temporal fusion within MSGNet, encompassing two branches: T-SGLM and TSB. The structure of HSTM is shown in Fig. \ref{net}.

In the T-SGLM, we segment the feature maps from two distinct moments, as derived from S-SGLM, into two separate sets of nodes. We then construct a temporal graph, following a methodology akin to that of S-SGLM. Subsequently, we deploy the sparse information aggregation mechanism to dynamically sift through and filter out redundant information between the feature graphs at adjacent moments.

\begin{figure}[t]  
	\centering
	\includegraphics[width=0.45\textwidth]{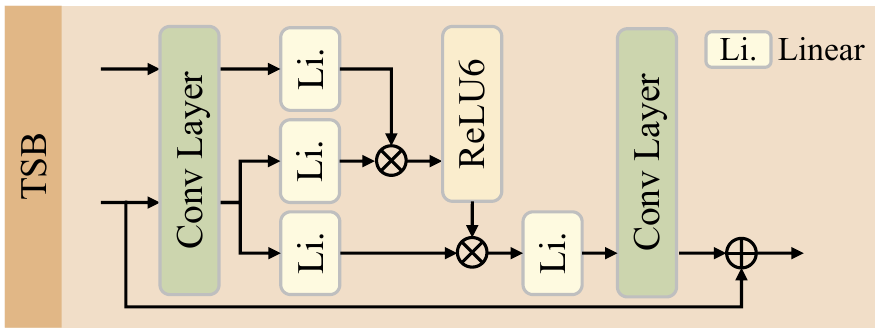}
	\caption{The structure of TSB. The fundamental components of the TSB are convolutional layers and linear layers. To enhance computational efficiency, we strategically intersperse element-wise multiplication operations, which rapidly project features into new high-dimensional subspaces. This design choice allows for swift and effective feature transformation while maintaining minimal computational overhead.} 
	\label{tsb}	
\end{figure}

However, a graph is an unordered data structure. Given that the spatial structural information conveyed by features may be overlooked during multimodal feature fusion and spatiotemporal information aggregation, we introduce a new branch to enhance the representation of similar spatial topological relationships across multiple frames. Inspired by the efficient design principles of StarNet, we propose the Temporal Star Block, the structure of which is illustrated in Fig. \ref{tsb}.
The TSB receives feature maps from two frames: the current frame and the previous frame. It is important to note that these feature maps have already fused the multimodal information from their respective moments. TSB initially employs convolutional and linear layers to encode the features, followed by element-wise multiplication to efficiently map them into high-dimensional feature subspaces, thereby facilitating the exploration of deep connections between the inputs. Finally, the decoding process is conducted through linear and convolutional layers. Additionally, skip connections are utilized to enhance the stability of the module.

\subsection{Detection Head}
In MSGNet, the detection head is responsible for further processing the multimodal spatiotemporal features after mid-stage fusion to generate the final prediction results. The detection head of MSGNet is a task decoupled design. Its main function is to generate the object category, location and confidence required for the object detection task through the feature map.

\begin{figure}[htbp]  
	\centering
	\includegraphics[width=0.48\textwidth]{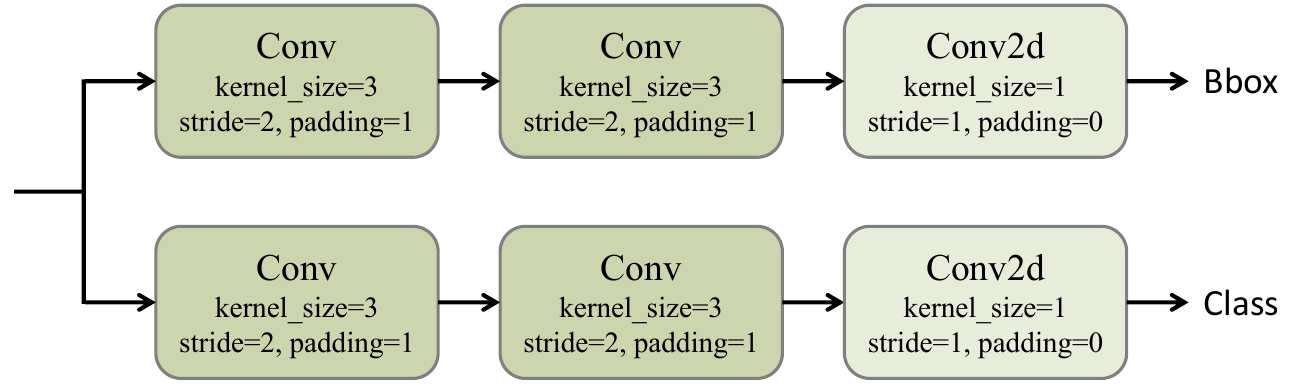}
	\caption{Schematic diagram of the detection head. The head employs a dual-branch decoupled architecture, characterized by symmetrical structural layouts yet independent learning processes for each branch. This design philosophy is specifically aimed at mitigating potential interferences between the two sub-tasks, thereby facilitating more efficient and targeted optimization of the localization and classification tasks.} 
	\label{head}	
\end{figure}

Fig. \ref{head} shows the specific structure of the detection head. It contains two branches with the same structure, each branch handles a subtask in object detection separately. The first branch is used to predict the position coordinates of the object and obtain multiple sets of quaternion coordinates to generate a set of predicted anchor boxes. The second branch is used to generate the object category and confidence predicted by the network. The reason why the two branches do not share learning parameters is to avoid the influence between the two subtasks.

\subsection{Loss Function}
Given that the RGBT VOD task is inherently a classic object detection problem, it encompasses two fundamental sub-tasks: object localization and object classification. Consequently, the loss function of MSGNet is designed to integrate these two components, thereby ensuring a comprehensive optimization framework for both sub-tasks.

Firstly, our object localization loss comprises two components: the Complete Intersection over Union (CIoU) loss \cite{zheng2020distance} and the Distribution Focal Loss (DFL) loss \cite{li2022generalized}. The CIoU loss is calculated as follows:
\begin{IEEEeqnarray}{c}
	L_{CIoU} = 1 - IoU + \frac{\rho^2(b,b^{gt})}{c^2} + {\alpha}v.
\end{IEEEeqnarray}
Here, $v$ is a parameter that quantifies the consistency of the aspect ratio between the predicted bounding box and the ground-truth box. The value of $\alpha$ is derived from $v$ and the IoU.  $\rho()$ denotes the Euclidean distance between the center points of the predicted box and the ground-truth box, represented by $b$ and $b^{gt}$, respectively. $c$ represents the diagonal length of the smallest enclosing rectangle that encompasses both the predicted box and the ground-truth box. And the definition of DFL loss is as follows:
\begin{IEEEeqnarray}{rCl}
\IEEEnonumber	L_{DFL}{(P_i,P_{i+1})}&=&-((y_{i+1}-y)log(P_i)\\
									&&+(y-y_i)log(P_{i+1})).
\end{IEEEeqnarray}
Here, $y$ represents the label. $y_i$ and $y_{i+1}$ denote the two predicted values nearest to $y$, and $P_i$ and $P_{i+1}$ represent their respective probabilities.
Secondly, the function employed to constrain the network's classification loss is the BCEWithLogitsLoss.
\begin{IEEEeqnarray}{rCl}
\IEEEnonumber	L_{BCE} &=&\frac{1}{N}{\sum}(y_i{\cdot}log({\sigma}(p_i))\\
		                               &&+(1-y_i){\cdot}log(1-{\sigma}(p_i))).
\end{IEEEeqnarray}
Here, $\sigma$ represents the Sigmoid function. $y_i$ is the classification label, and $p_i$ is the probability of predicting each category.

Therefore, the total loss of the MSGNet is composed as follows:
\begin{IEEEeqnarray}{c}
	L_{Total} = L_{CIoU}+L_{DFL}+L_{BCE}.
\end{IEEEeqnarray}
The total loss function serves to guide and optimize the learning process of MSGNet in its entirety.

\section{Experiments}
This section introduces the experimental background, demonstrates the effectiveness of our proposed method through rigorous testing, provides insights that clarify the rationale behind our design choices, and carries out analyses based on these findings.
\subsection{Experimental Configuration}
\begin{table*}[t]
	\centering
	\caption{
		We evaluate MSGNet alongside current mainstream detection models on UVT-VOD2024 and VT-VOD50, emphasizing the best results in \textbf{bold}. The symbol ``-'' indicates that the measurement conditions were not met or that the result remains unobtainable. Additionally, the symbol ``$\dagger$'' denotes instances in which the network exhibits convergence difficulties, necessitating an increase in both the depth and width of the architecture. At this stage, discrepancies may arise between the reported number of parameters and computational load relative to the data presented in this table.
	}
	\setlength{\tabcolsep}{0.5mm}{
		\begin{tabular}{lcc|cc|cc|ccc}
			\toprule
			\multirow{2}[2]{*}{Methods} & \multirow{2}[2]{*}{Backbone} & \multirow{2}[2]{*}{Type} & \multicolumn{2}{c|}{\makecell[c]{UVT-VOD2024\\(alignment-free)}} & \multicolumn{2}{c|}{\makecell[c]{VT-VOD50\\(aligned)}} & \multirow{2}[2]{*}{FPS} & \multirow{2}[2]{*}{Params(M)} & \multirow{2}[2]{*}{FLOPs(G)} \\
			&       &       & AP50(\%) & AP(\%) & AP50(\%) & AP(\%) &       &       &  \\
			
			\midrule
			YOLOV3 \cite{redmon2018yolov3}& Darknet53 & Image    &25.6   &13.5 &33.9& 17.4& 69.9 & 103.7 &283\\
			
			YOLOV5\_M \cite{yolov5} & CSPDarknet53 & Image    &  23.9 & 12.5&-&-& \textbf{294.1} & 25.1 & 64.4 \\
			
			CFT \cite{qingyun2021cross} & CFB & Image    &  6.7 &  2.4 &42.5&18.9& 222.2 & 73.7 & - \\
			
			YOLOX\_L \cite{ge2021yolox}& Darknet53 & Image     &  16.3   & - &-&-&104.8 &54.2 & 155.8 \\
			
			YOLOV6\_M \cite{li2022yolov6}  & EfficientRep & Image    & 22.7  & 12.1 &-&- & 169.5& 52 & 161.6 \\
			
			YOLOV7 \cite{10204762} & CSPDarknet53 & Image   &23 &10.4 &37.7&16.5& \textbf{294.1} & 36.5& 103.3 \\
			
			YOLOV9-C \cite{wang2024yolov9}&GELAN&Imgae& 27.3&14.5&49.1&26.9&99&25.5&103.7\\
			
			YOLOV10-M \cite{wang2024yolov10}&CSPNet&Image&17.1&8.7&46.2&25.2&210&16.5&64\\
			Efficientdet \cite{tan2020efficientdet} &EfficientNet&Image&20.2&8.8&-&-&87&20.0&100\\
			
			TOOD \cite{feng2021tood} &ResNet-50&Image&15.9&7.3&36.3&19&25.8&32&199\\
			
			Deformable DETR \cite{zhu2021deformable}&ResNet-50&Imgae&7.7&2.9&42.5&23.3&20.7&41.1&197\\
			
			RT-DETR \cite{zhao2024detrs}&ResNet-50&Imgae&17&7.9&40.2&21.6&-&42.7&130.5\\
			
			DINO \cite{zhang2022dino}&ResNet-50&Image&29.4&13.7&47.4&25.9&16.7&47.7&274\\
			
			AlignDETR \cite{cai2023align}&ResNet-50&Image&21.1&9.2&-&-&12.9&47.5&235\\
			
			DDQ DETR \cite{zhang2023dense} & ResNet-50 & Image     & 21.1  & 9.1 &48.3&26.5& 13 & 48.3& 275 \\
			
			DiffusionDet \cite{chen2023diffusiondet}&ResNet50&Image&21.4&9.6&46.9&25.1&-&-&-\\
			\midrule
			DFF \cite{zhu2017deep}  & ResNet-50 & Video  & 9.2  & 3.9 &33.5&14.1 & 40.4 & 62.1 & \textbf{24.9} \\
			
			FGFA \cite{zhu2017flow}  & ResNet-50 & Video   & 16.7 & -  &35.1&15.8& 9 & 64.5& 41 \\
			
			RDN \cite{deng2019relation} & ResNet-50 & Video   &  16.9  &  -  &40&-& 11.3 &    -   & - \\
			
			SELSA \cite{wu2019sequence}  & ResNet-50 & Video   &  12.6 &  4.6&39.4&17.4 & 10.5&  -     & - \\
			
			MEGA \cite{chen2020memory}& ResNet-50 & Video   & 15.4  & -    &27.8&-& 16.2 &   -    & - \\
			
			Temporal ROI Align \cite{gong2021temporal} & ResNet-50 & Video   &  11.1   & 3.9  &38&17& 5.1 &    -   & - \\
			
			CVA-Net \cite{10.1007/978-3-031-16437-8_59}  & ResNet-50 & Video &  16.4    & 6.4 &39.7&19.7 & 6.9 & 41.6&548.1 \\
			
			STNet \cite{qin2023spatial}  & ResNet-50 & Video     & 15.7  & 6.5 &38.4&18.4& 5 & 41.6 & 752.3 \\
			
			EINet \cite{tu2023erasure}  & Darknet53 & Video    & 20.7& -  &46.3&24& 204.2 & 11.6& 78.2 \\
			PTMNet \cite{wang2025high}  & CSPDarknet53 & Video    & -& -  &51.4&27.1& 72.5 & 11.4& 77.8 \\
			MDLNet-S \cite{wang2024unveiling}  & CSPDarknet53 & Video    & 26.9& 13.5  &54.4&30.2& 123.5 & 22.7& 69.2 \\
			MDLNet-L \cite{wang2024unveiling}  & CSPDarknet53 & Video    & 31.8& 15.5  &\textbf{57.9}&\textbf{32.5}& 54.6 & 89.7& 271.4 \\
			\midrule
			MSGNet (Ours) & CSPDarknet53 &Video   & \textbf{38.5}   & \textbf{20.7} & 55.6$^{\dagger}
			$& 31.8 & 232.6 &\textbf{6.58} & 74.69 \\
			\bottomrule
	\end{tabular}}
	
	\label{compara1}
\end{table*}
\subsubsection{Environment}
Regarding hardware, we utilize two NVIDIA GeForce RTX 3090 GPUs to enhance graphics processing capabilities. On the software front, we have implemented the Ubuntu 22.04.2 operating system. For the training phase, the PyTorch deep learning framework, version 2.1.1, is employed, while coding is conducted in Python version 3.10.13.
\subsubsection{Network Setup}
The network training begins with a complete initialization and spans 100 training epochs. The network is designed to accept images of size 640 $\times$ 640, with a batch size of 18 images per GPU (Graphics Processing Unit). The Non-Maximum Suppression (NMS) Intersection over Union (IOU) threshold is set at 0.65, and the initial learning rate is fixed at 0.01. The optimizer’s momentum is configured to 0.938. Data augmentation techniques include color space adjustments and basic horizontal mirroring only. More advanced augmentation methods, such as mosaic \cite{bochkovskiy2020yolov4}, are intentionally disabled to preserve the coherence of multimodal data.

\subsubsection{Datasets}
The datasets employed to evaluate detector performance are VT-VOD50 \cite{tu2023erasure} and UVT-VOD2024 \cite{wang2024unveiling}. VT-VOD50 represents the first benchmark established for RGBT Video Object Detection (VOD) and has undergone manual data preprocessing to ensure the spatial alignment of RGBT image pairs at the image level. In contrast, UVT-VOD2024 is a larger-scale dataset intended for alignment-free RGBT VOD. Relative to VT-VOD50, it encompasses a wider variety of real-world scenes, object categories, and a larger number of videos. Notably, this dataset does not undergo any additional preprocessing beyond annotation; instead, it retains the original spatially alignment-free multimodal images. This characteristic is particularly relevant for real-world application scenarios. As certain methods utilize single-modal frameworks, our approach involves performing pixel-wise addition of multimodal images followed by inputting them into these networks.
\subsubsection{Metrics}
To assess the performance of each model, we employ standard metrics in the field of object detection. Specifically, we consider the number of model parameters and computational load, followed by Average Precision (AP) and AP50 for accuracy, in addition to Frames Per Second (FPS) for inference speed.

\subsection{Comparative Experiments}
Building upon the experimental design outlined above, we undertake a comprehensive evaluation of MSGNet using a substantial number of image-based and video-based detectors, with the results presented in Tab. \ref{compara1}.

Initial observations reveal that MSGNet significantly outperforms the comparative methods in detection accuracy on UVT-VOD2024, achieving a relative improvement of 30.9\% over the second-best method, DINO \cite{zhang2022dino} while operating over ten times faster. In terms of inference speed, MSGNet ranks just below YOLOV5\_M \cite{yolov5} and YOLOV7 \cite{10204762}, achieving 232.6 FPS. Notably, these two detectors accept only a single RGB image as input, resulting in detection accuracy that is approximately half that of our network. Furthermore, MSGNet exhibits a clear advantage in parameter count compared to all comparison methods. Although the graph computations involved in our approach lead to a slightly higher computational load than certain end-to-end networks, such as MDLNet-S \cite{wang2024unveiling} and YOLOV5\_M \cite{yolov5}, it remains significantly more efficient than most detectors, including CVA-Net \cite{10.1007/978-3-031-16437-8_59} and STNet \cite{qin2023spatial}.

Next, we assess the performance of all methods on the spatially aligned dataset VT-VOD50. Our MSGNet demonstrates significant advantages in both speed and accuracy compared to existing methods. This superiority can be attributed to its one-stage detection strategy and compact model size. Specifically, MSGNet achieves an AP50 score that is 13.2\% higher than the suboptimal image-based method, YOLOV9-C \cite{wang2024yolov9}, and 5.3\% higher than the suboptimal video-based method, MDLNet-S \cite{wang2024unveiling}, in terms of overall AP. Additionally, the detection delay is approximately half that of the aforementioned methods.

\begin{table*}[t]
	\centering
		\caption{Results of the ablation experiments on UVT-VOD2024. HSTM includes T-SGLM and TSB modules here.}
	\begin{tabular}{cccccccccc}
		\toprule
		Groups & Type  & HSTM & S-SGLM &APL & AP50(\%) & AP(\%) & FPS & Params(M) & FLOPs(G) \\
		\midrule
		a (baseline)     & RGB   &       &       &       & 22.7  & 10.3  & 1111.1 & 3.01  & 8.2 \\
		b     & RGB   & \Checkmark     &       &       & 24.3  & 12.8  & 555.6 & 5.3   & 68.2 \\
		c     & Thermal     &       &       &       &  3.91     &  1.53     & 1111.1 & 3.01  & 8.2 \\
		d     & Thermal     & \Checkmark     &       &       &   4.15    &   1.52    & 555.6 & 5.3   & 68.2 \\
		e     & RGBT  &       &     &       & 22.4  & 10.6  & 692.5 & 3.9   & 9.4 \\
		f     & RGBT  &       & \Checkmark     &       & 25.1  & 11.6  & 555.6 & 4.2   & 11.3 \\
		g     & RGBT  &       & \Checkmark     & \Checkmark     & 34    & 17.1  & 419.7 & 4.29  & 11.4 \\
		h     & RGBT  & \Checkmark     & \Checkmark     & \Checkmark     & 38.5  & 20.7  & 232.6 & 6.58  & 74.69 \\
		\bottomrule
	\end{tabular}

	\label{ablation}
\end{table*}
We analyze the information presented in Tab. \ref{compara1} from several perspectives. Firstly, when viewed horizontally, it is apparent that the detection accuracy of existing methods on UVT-VOD2024 significantly lags behind that on VT-VOD50. The Average Precision at IoU 50 (AP50) for most approaches in the former dataset is approximately half that of the latter. This disparity suggests that the spatially alignment-free multimodal information inherent in UVT-VOD2024 presents substantial challenges for network learning, even though this scenario is more indicative of real-world applications. In contrast, our MSGNet achieves an AP50 on UVT-VOD2024 that is only about 10\% lower than that recorded on VT-VOD50, thus demonstrating its effectiveness.
From a vertical perspective, while most existing detectors exhibit similar performance to MSGNet on the VT-VOD50 dataset, their performance on UVT-VOD2024 is markedly inferior. These findings collectively indicate that our proposed graph representation learning effectively addresses the challenges posed by spatially alignment-free RGBT image pairs.

\subsection{Ablation Experiments}
\subsubsection{Effectiveness Validation of Designed Modules}
To validate the practical effectiveness of each component in our design, we conducted a comprehensive set of ablation experiments, the results of which are detailed in Tab. \ref{ablation}. Comparing groups a and b, as well as groups c and d, it becomes clear that HSTM demonstrates robust performance across both modalities, adeptly capturing continuous dependencies in both temporal and spatial contexts. However, the results for group c are significantly lower than those for group a. This performance discrepancy stems from the fact that annotations were conducted exclusively on RGB images. When training relies solely on thermal images, variations in spatial resolution can result in mismatches between the labels and images, ultimately leading to degraded performance. Group e illustrates the results utilizing RGBT data without any modules. Groups f and g build upon E by incorporating S-SGLM and APL, respectively, demonstrating notable enhancements in performance. Particularly, the improvement attributed to APL underscores the importance and efficacy of initial coarse alignment. Group h showcases the substantial improvement in detection performance achieved by incorporating all designed modules into the baseline configuration.

\subsubsection{The Impact of Varying Sparse Settings on MSGNet}
To assess the efficacy of robust sparse information aggregation within S-SGLM, we conduct the experiments detailed in Tab. \ref{threshold}.
Edges in the multimodal graph with weights below the designated threshold are deemed redundant and eliminated. It is observed that both fully dense connections (threshold of 0) and excessively sparse connections (threshold of 0.75) lead to a significant decrease in network accuracy. Optimal performance is attained when a threshold of 0.25 is applied to remove superfluous connection edges.
\begin{table}[htbp]
	\centering
		\caption{The impact of different thresholds on network performance. We highlight the best results in \textbf{bold}. Here, ``$\ddagger$" symbolizes the configuration employed by our proposed method.}
	\begin{tabular}{ccc}
		\toprule
		Threshold & AP50(\%) & AP(\%) \\
		\midrule
		0&30.3&15.2\\
		$0.25^{\ddagger}$  & \textbf{34}    & 17.1 \\
		0.5   & 32.9  & \textbf{17.2} \\
		0.75  & 30.1  & 13.7 \\
		\bottomrule
	\end{tabular}

	\label{threshold}
\end{table}

In addition to using a set threshold to filter out connections with too low weights, we also select the top K pairs of nodes from all connection sets in order of weight to balance the performance and efficiency of the network. We conduct experiments on T-SGLM and S-SGLM using different K values, and the results are shown in Tab. \ref{top_k1} and Tab. \ref{top_k2} respectively. GPU Mem (Memory) refers to the memory occupied by the model on a single GPU.

\begin{table}[htbp]
	\centering
		\caption{Experimental results using different K values in T-SGLM. Here, ``$\ddagger$" symbolizes the configuration employed by our proposed method.}
	\begin{tabular}{cccc}
		\toprule
		K & AP50(\%) & AP(\%) &GPU Mem (G)\\
		\midrule
		10&31.9&17.9&11.8\\
		25  & 37.2   & 19.8 &12.3\\
		50   & 28.4 & 15.5 &13 \\
		75& 38.3&19.9&14.4\\
		$100^{\ddagger}$& \textbf{38.5}    & \textbf{20.7} &16.2\\
		125&34.2&17&17\\
		\bottomrule
	\end{tabular}

	\label{top_k1}
\end{table}

\begin{table}[htbp]
	\centering
		\caption{Experimental results using different K values in S-SGLM. Here, ``$\ddagger$" symbolizes the configuration employed by our proposed method.}
	\begin{tabular}{cccc}
		\toprule
		K & AP50(\%) & AP(\%) &GPU Mem (G)\\
		\midrule
		10&30.9&17.5&15.3\\
		$25^{\ddagger}$  & \textbf{38.5}    & \textbf{20.7} &16.2\\
		50   & 38.3  & 20.1&17.1 \\
		75&34.2 &17.5&18.9\\
		100&35&19.3&20.8\\
		125&32.2&16.5&23.5\\
		\bottomrule
	\end{tabular}

	\label{top_k2}
\end{table}

It can be seen that when the K value increases, the computational cost of the network will also increase, which is reflected in the need to occupy more GPU memory during network training. At the same time, it will also reduce the efficiency of network training and inference. In S-SGLM, only the first 25 pairs of nodes with large weights are needed to achieve the best network performance. In T-SGLM, the network achieves the highest accuracy only when the value of K is increased to 100. We believe that this is because in S-SGLM, both sides of each pair of nodes come from feature maps of different modalities, and the feature domains are quite different, resulting in less available information. In T-SGLM, both sides of each pair of nodes contain inherent RGB information, and the difference in feature space is reduced, so more information can be aggregated, resulting in the need for more node pairs to achieve optimal performance.

\subsubsection{The Impact of Different Input Structures on APL}
To further investigate the structural advantages of APL, we conducted a series of ablation experiments focusing on key input combination structures. The results are presented in Tab. \ref{Aplinput}. The network demonstrates optimal performance when the two inputs are concatenated along the channel dimension. Following closely in performance is the scenario where the two inputs are combined using element-wise addition. However, a notable decline in detector performance is observed when RGBT features are multiplied element-wise. This reduction in performance can be attributed to the divergence of feature maps from their original distribution, which impairs the network's ability to accurately discern the initial spatial distribution differences.
\begin{table}[htbp]
	\centering
	\caption{
		The impact of varying input structures on APL's performance.}
	\setlength{\tabcolsep}{1.1mm}{
		\begin{tabular}{cccc}
			\toprule
			Gropus& Methods & AP50(\%) & AP(\%) \\
			\midrule
			a & Channel Concatenation &34 & 17.1\\
			b  & Element-wise Addition&33.9    & 17 \\
			c   &Element-wise Multiplication &30.5  & 15.6 \\
			\bottomrule
	\end{tabular}}
	
	\label{Aplinput}
\end{table}

\begin{figure*}[t]
	\centering
		\subfloat[]
		{\includegraphics[width=0.4\linewidth]{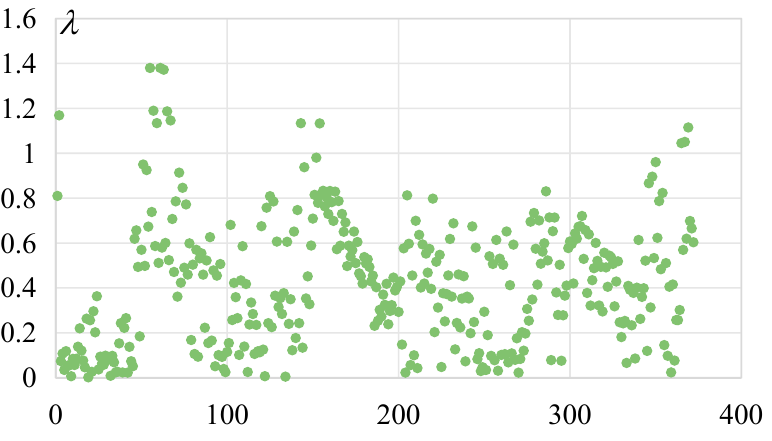}
		\label{apl1}}
		\subfloat[]
		{\includegraphics[width=0.6\linewidth]{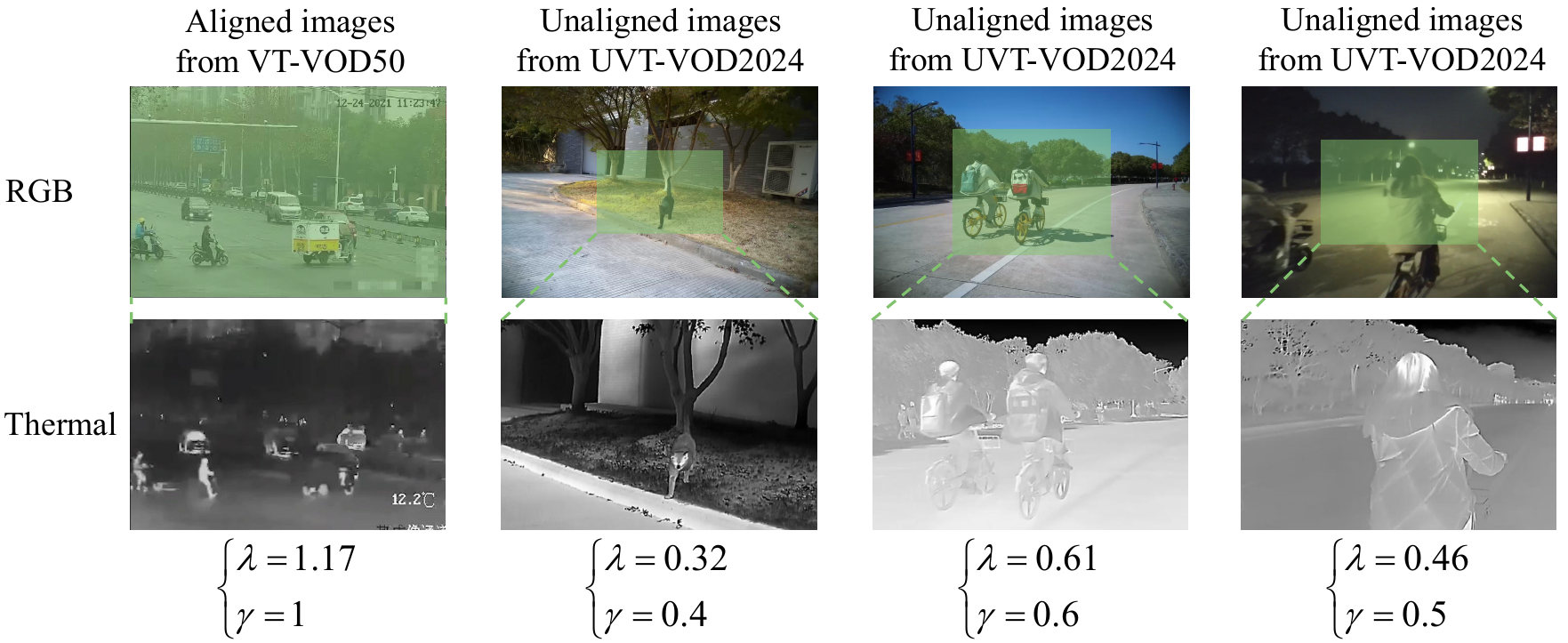}\label{apl2}}

	\caption{Visualization of APL Experimental Results. The figure (a) represents the distribution of predicted values for $\lambda$ in the test set of UVT-VOD2024. The figure (b) shows the value of $\lambda$ and $\gamma$ predicted by APL in different scenarios during the test phase.}
	\label{apl}
\end{figure*}

\subsubsection{Verification of the Effectiveness of T-SGLM and TSB in HSTM}
We conduct ablation experiments on the two components of HSTM, with the results presented in Tab. \ref{HSTM}. The findings in Tab. \ref{HSTM} are based on training with only $Frame^{RGB}_{t-1}$ and $Frame^{RGB}_{t}$. When T-SGLM and TSB are deployed individually, the network achieves similar accuracy improvements. However, when both methods are combined, the AP50 index exhibits an additional 1\% enhancement. This result demonstrates how T-SGLM's filtering of temporal redundancy complements TSB's modeling of local structural information.
\begin{table}[htbp]
	\centering
		\caption{
		Performance comparison of various method configurations in HSTM.}
	\setlength{\tabcolsep}{1.3mm}{
		\begin{tabular}{cccc}
			\toprule
			Methods & AP50(\%) & AP(\%) & Params(M)  \\
			\midrule
			w/o TSB \& T-SGLM & 22.7  & 10.3    & 3.01   \\
			w/ T-SGLM & 23.3  & 11    & 4.74   \\
			w/ TSB & 23.4  & 11.3  & 3.58  \\
			w/ TSB \& T-SGLM & 24.3  & 12.8  & 5.3    \\
			\bottomrule
	\end{tabular}}

	\label{HSTM}
\end{table}

\subsection{The Actual Effect of APL}

Does APL accurately capture the spatial correspondence of multimodal data distributions? To investigate this question, we conducted experiments and analyses. Initially, we tested MSGNet on the UVT-VOD2024 dataset, visualizing the predictions made by APL through a scatter plot, as shown in Fig. \ref{apl1}. The results indicate that the majority of points are distributed between 0 and 0.8, demonstrating that the network effectively recognizes significant differences in spatial distribution between RGB and thermal images. A smaller number of points are situated between 0.8 and 1.4, suggesting that the network determines this portion of the RGB image does not require cropping, as the additional scene information from the RGB data remains pertinent for current object detection.

We select several groups of images from the VT-VOD50 and UVT-VOD2024 datasets for testing, recording their corresponding values of $\lambda$ and $\gamma$, as illustrated in Fig. \ref{apl2}. In the first set of spatially aligned images from VT-VOD50, APL predicted a value of 1.17 for $\lambda$, with a discrete mapping of 1 for $\gamma$. This finding indicates that no cropping is needed, which aligns with the actual conditions since there are no discrepancies in spatial distribution. In contrast, the second set of alignment-free RGBT image pairs from UVT-VOD2024 produced a predicted value of 0.32 for $\lambda$ and a cropping ratio of 0.4 for $\gamma$. Notably, this set features a single predefined object—a cat—suggesting that the network should predominantly focus on the area surrounding the cat, an inference supported by the cropping ratio of 0.4. Similarly, the subsequent two groups revealed distinct predicted $\gamma$ values, consistent with the actual characteristics of the image pairs.

\subsection{Choice of $\lambda$ and $\gamma$ in APL}
In APL, why do we need to set artificial rules for the model-predicted $\lambda$ to further generate $\gamma$? We will analyze and explain it in this section. 

As can be seen from Fig. \ref{apl1}, the predicted values of $\lambda$ are distributed between 0 and 1.5, and most of them are distributed below 0.8. a determines the scale of the RGB feature map to interact with the thermal image feature map, so in theory it should be less than or equal to 1. This means that we need to perform secondary processing on $\lambda$.

At the same time, we believe that the area of the RGB feature map used for interaction is not strictly continuous, that is, it should not be strictly spatially corresponding to the thermal feature map. For example, let's look at the second column of Fig. \ref{apl2}. The predefined object in the image is only a cat. This cat is just near the center of the RGB image and occupies a small proportion. Then we can assume that the network predicts a small $\lambda$ value such as 0.1 for this image when it first starts learning. However, compared with the thermal image, we can see that the spatial proportion of the thermal image in the RGB image is more than 0.1. In other words, we also need to take into account the spatiotemporal information around the object. So we use the rules we set to map this predicted value $\lambda$ to $\gamma$. When $\lambda$ is less than 1, $\gamma$ is greater than or equal to $\lambda$. When $\lambda$ is less than 1, we choose to divide the interval from 0 to 1 where a falls into 5 areas when setting the rules, corresponding to different $\gamma$. In order to verify the above analysis, we also conduct a set of experiments, using $\lambda$ and $\gamma$ in APL to complete the learning of the network, and the results are shown in Tab. \ref{aorb}.

\begin{table}[htbp]
	\centering
		\caption{Use $\lambda$ and $\gamma$ respectively to determine different results for the interaction area in the RGB image.}
	\begin{tabular}{cccc}
		\toprule
		$\lambda$ & $\gamma$ & AP50(\%) & AP(\%)\\
		\midrule
		\checkmark & & 33.4&18.2\\
		&\checkmark & \textbf{38.5}&\textbf{20.7}\\
		\bottomrule
	\end{tabular}

	\label{aorb}
\end{table}
\begin{figure*}[t]  
	\centering
	\includegraphics[width=0.9\textwidth]{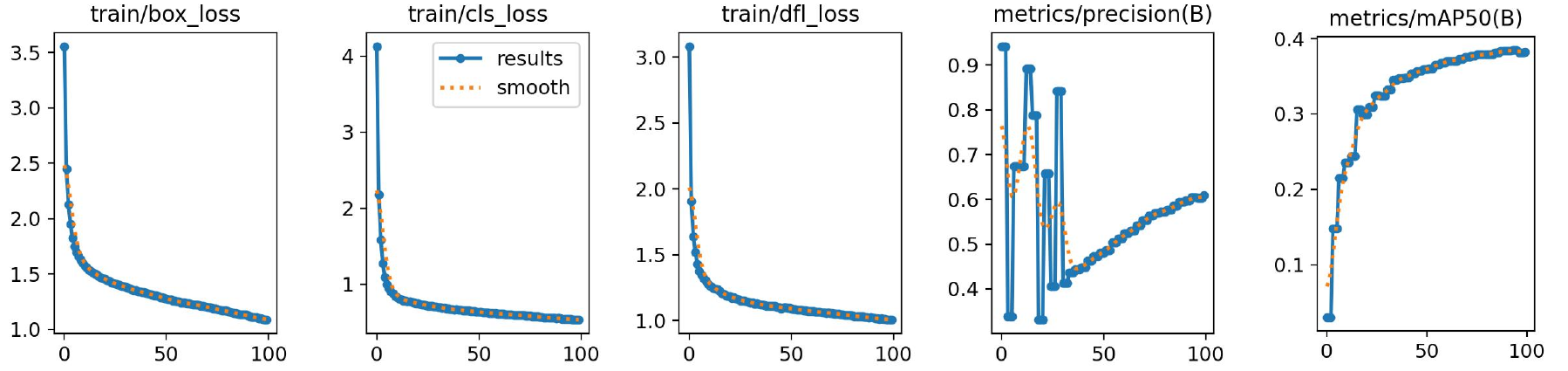}
	\caption{During the training of MSGNet, we record and plot the changes in the loss function, precision and AP50. Among these metrics, box\_loss and dfl\_loss reflect the MSGNet's performance in object localization, while cls\_loss represents the network's performance in object classification. Precision denotes the probability that a sample predicted by MSGNet is correctly identified. Meanwhile, AP50 signifies the detection accuracy of MSGNet evaluated on the test set under the condition of an IoU threshold of 0.5.}  
	\label{loss}	
\end{figure*}
The results show that the best results are achieved when we use our predefined rules to obtain $\gamma$ and then perform region selection on the RGB feature map.

As mentioned above, we divide the area from 0 to 1 into 5 groups to map $\lambda$ to $\gamma$. In order to explore the impact of different groupings on network performance, we construct a set of experiments, which are shown in Tab. \ref{area}. The experiment confirms the effectiveness of using $\gamma$ to help network learning.

\begin{table}[htbp]
	\centering
		\caption{The impact of different ways of mapping $\lambda$ to $\gamma$ on network performance. Here, ``$\ddagger$" symbolizes the configuration employed by our proposed method.}
	\begin{tabular}{ccc}
		\toprule
		Number of groups &  AP50(\%) & AP(\%)\\
		\midrule
		2&30.1&15.9\\
		3&38.3&19.8\\
		4&31.7&16.2\\
		$5^{\ddagger}$ &\textbf{38.5}&\textbf{20.7}\\
		6&35.9&18.2\\
		\bottomrule
	\end{tabular}

	\label{area}
\end{table}

However, non-continuous $\gamma$ values will make it difficult for the network to learn. Specifically, when training the network, there will be requirements for batch size and network width, otherwise the network learning will not converge. Our future work will consider improving this aspect.

\subsection{Training Log}
During the training process of MSGNet, we meticulously document the temporal evolution of various data metrics, as illustrated in Fig. \ref{loss}. Specifically, the results highlight the reduction trajectories of both localization and classification losses. Notably, the network's training loss exhibits a rapid decline within the initial five training epochs, a phenomenon attributable to the implementation of a warm-up learning rate strategy. This strategy effectively facilitates the model's adaptation to the training data, thereby expediting the convergence of the loss function.
In the subsequent training epochs, the decline in these losses gradually stabilizes, indicating that the network has reached a phase of more nuanced optimization. Additionally, we also assess the model's performance using precision, which is defined as the proportion of true positives correctly identified by the evaluation model. Following the initial phase of learning, the network's accuracy stabilizes around 60\% in the later stages of training. This suggests that, while the network demonstrates satisfactory performance, there remains significant potential for further enhancement of its detection capabilities.
Lastly, we present the AP50 metric as an indicator of the network's overall detection accuracy. After completing 100 training epochs, the AP50 score of MSGNet converges to 38.5. This value serves as a critical benchmark, reflecting the network's capacity for object detection and highlighting the need for continued refinement to achieve higher accuracy levels.

\section{Future Directions}
Many existing RGBT fusion methods operate independently and are specifically tailored for different alignment scenarios. Some methods are designed for perfectly aligned RGBT image pairs, while others excel at weakly aligned pairs at the pixel level or focus solely on scenarios with significant misalignment. However, due to the barriers between design and training data, these models often struggle to generalize effectively to other scenarios.

Our proposed MSGNet aims to unify the perception of various scenarios, including aligned, weakly aligned, and misaligned RGB-T image pairs. Although we have demonstrated its capability in this paper, we found that network training becomes difficult to converge when the model depth increases. This issue may be caused by the discontinuous value of $\gamma$ we set, which affects the smoothness of the feature alignment process. Therefore, in future research, we will develop more robust detectors that can handle such challenges. We will even consider extreme cases where the view angle of Thermal images is larger than that of RGB images. By exploring these innovations, we aim to enhance the adaptability and robustness of RGBT fusion methods in various real-world scenarios.

\section{Conclusion}
To overcome the limitations of existing RGBT VOD methods, which predominantly rely on precisely aligned image pairs, we introduce MSGNet, a novel architecture designed for alignment-free RGBT fusion. Specifically, our approach addresses the challenges associated with spatial mismatches between RGB and Thermal modalities. First, we incorporate the APL to perceive and mitigate the significant scale discrepancies in spatial distribution between RGB and Thermal images. This module enables the network to adaptively align the features of the two modalities, effectively reducing the impact of misalignment on the fusion process. Second, the S-SGLM is introduced to refine the fusion process. This module robustly integrates highly correlated thermal features into RGB features by leveraging semantic relationships and contextual information. By doing so, it enhances the representation of objects in the RGB domain with the complementary thermal information. Furthermore, to address redundancy within the temporal domain, the T-SGLM filters out unnecessary information between features at adjacent moments. This module ensures that only relevant temporal information is retained, thereby improving the efficiency and accuracy of the fusion process. Additionally, the TSB is designed to maintain the temporal consistency of local spatial structures across frames, ensuring that the network captures coherent and meaningful motion patterns. Overall, the proposed MSGNet is highly adaptable to RGBT image pairs with varying spatial distribution differences, making it a versatile solution for diverse real-world scenarios. Our approach not only enhances the robustness and accuracy of RGBT VOD but also paves the way for the future unification of RGBT fusion techniques across different spatial and temporal configurations.

	\bibliographystyle{IEEEtran}
	\bibliography{IEEEabrv,main}
\end{document}